\documentclass[sigconf]{acmart}
\AtBeginDocument{%
  \providecommand\BibTeX{{%
    \normalfont B\kern-0.5em{\scshape i\kern-0.25em b}\kern-0.8em\TeX}}}

\copyrightyear{2022}
\acmYear{2022}
\setcopyright{acmcopyright}\acmConference[MM '22]{Proceedings of the 30th ACM International Conference on Multimedia}{October 10--14, 2022}{Lisboa, Portugal}
\acmBooktitle{Proceedings of the 30th ACM International Conference on Multimedia (MM '22), October 10--14, 2022, Lisboa, Portugal}
\acmPrice{15.00}
\acmDOI{10.1145/3503161.3548262}
\acmISBN{978-1-4503-9203-7/22/10}

\acmSubmissionID{2178}

\usepackage{graphicx}
\usepackage{amsmath}
 
\usepackage{amssymb}
\usepackage{booktabs}
\usepackage{bbding}
\usepackage{color,xcolor}
\usepackage{multirow}
\usepackage{float}
\usepackage{adjustbox}
\usepackage{url}
\usepackage{hyperref}
\usepackage{subcaption}

\newcommand{\reflabel}{dummy} 

\newcommand{\seclabel}[1]{\label{sec:\reflabel-#1}}
\newcommand{\secref}[2][\reflabel]{Section~\ref{sec:#1-#2}}

\renewcommand{\eqref}[2][\reflabel]{(\ref{eq:#1-#2})}

\newcommand{\etal}{et al.}
\newcommand{\eg}{e.g.}
\newcommand{\ie}{i.e. }

\newcommand{\R}{\mathbb{R}}

\newcommand{\pcl}{X} 

\newcommand{\jointloc}{J}
\newcommand{\vertloc}{V}
\newcommand{\mesh}{\mathcal{M}} 
\newcommand{\joint}{\mathcal{J}} 
\newcommand{\trans}{\gamma} 
\newcommand{\loss}{\mathcal{L}} 
\newcommand{\smpl}{\theta} 
\newcommand{\pose}{\alpha} 
\newcommand{\shape}{\beta}

\newcommand{\cajjoint}{J_{t}} 
\newcommand{\cajvertices}{V_{t}}

\usepackage[capitalize]{cleveref}
\crefname{section}{Sec.}{Secs.}
\Crefname{section}{Section}{Sections}
\Crefname{table}{Table}{Tables}
\crefname{table}{Tab.}{Tabs.}

\begin{document}

\title{mmBody Benchmark: 3D Body Reconstruction Dataset and Analysis for Millimeter Wave Radar}

\author{Anjun Chen}
\affiliation{
    \institution{Zhejiang University}
    \city{Hangzhou}
    \country{China}
}
\authornote{Equal contribution.}
\author{Xiangyu Wang}
\affiliation{
    \institution{Zhejiang University}
    \city{Hangzhou}
    \country{China}
}
\authornotemark[1]
\author{Shaohao Zhu}
\affiliation{
    \institution{Zhejiang University}
    \city{Hangzhou}
    \country{China}
}
\author{Yanxu Li}
\affiliation{
    \institution{Zhejiang University}
    \city{Hangzhou}
    \country{China}
}
\author{Jiming Chen}
\affiliation{
    \institution{Zhejiang University}
    \city{Hangzhou}
    \country{China}
}
\author{Qi Ye}
\affiliation{
    \institution{Zhejiang University}
    \city{Hangzhou}
    \country{China}
}

\authornote{Corresponding Author Qi Ye. Qi Ye is with the College of Control Science and Engineering, the State Key Laboratory of Industrial Control Technology, Zhejiang University, and also with the Key Laboratory of Collaborative sensing and autonomous unmanned systems of Zhejiang Province. Email: qi.ye@zju.edu.cn}

\renewcommand{\shortauthors}{Anjun Chen, et al.}
\begin{abstract}

Millimeter Wave (mmWave) Radar is gaining popularity as it can work in adverse environments like smoke, rain, snow, poor lighting, etc. Prior work has explored the possibility of reconstructing 3D skeletons or meshes from the noisy and sparse mmWave Radar signals. However, it is unclear how accurately we can reconstruct the 3D body from the mmWave signals across scenes and how it performs compared with cameras, which are important aspects needed to be considered when either using mmWave radars alone or combining them with cameras. To answer these questions, an automatic 3D body annotation system is first designed and built up with multiple sensors to collect a large-scale dataset. The dataset consists of synchronized and calibrated mmWave radar point clouds and RGB(D) images in different scenes and skeleton/mesh annotations for humans in the scenes. With this dataset, we train state-of-the-art methods with inputs from different sensors and test them in various scenarios. The results demonstrate that 1) despite the noise and sparsity of the generated point clouds, the mmWave radar can achieve better reconstruction accuracy than the RGB camera but worse than the depth camera; 2) the reconstruction from the mmWave radar is affected by adverse weather conditions moderately while the RGB(D) camera is severely affected. Further, analysis of the dataset and the results shadow insights on improving the reconstruction from the mmWave radar and the combination of signals from different sensors.

\end{abstract}

\ccsdesc[500]{Computing methodologies~Reconstruction}
\ccsdesc[300]{Computing methodologies~Point-based models}
\ccsdesc[100]{Computing methodologies~Motion capture}

\keywords{Millimeter wave radar, 3D body reconstruction, Dataset, Analysis}



\maketitle
\section{Introduction}
\label{sec:intro}

Recently, wireless sensing is increasingly gaining popularity in areas like autonomous driving  \cite{wang2021rethinking,qian2021robust}, human activity detection  \cite{li2019making, singh2019radhar}, automotive electronics  \cite{qian2021robust, li2021real}, and UAV  \cite{Axelsson_2021_CVPR, vsipovs2020lightweight} due to its capability to work in low-visibility environments such as dense fog, smoke, snow-storm, rain, etc. \cite{garcia2018robust}, as well as less privacy leakage compared with RGB(D) cameras  \cite{wang2019person}. 

The applications of wireless sensing have been explored in SLAM  \cite{aladsani2019leveraging, lu2020milliego}, vehicles detection and location  \cite{bijelic2020seeing, qian2021robust}, car imaging  \cite{guan2020through}, human detection and pose estimation \cite{zhao2018through, zhao2018rf, wang2019person, li2019making}, vital signs monitoring \cite{adib2015smart, hsu2017zero}, etc. Despite the diversity of those applications, works on human body reconstruction from wireless signals are still limited. The main reason lies in the low spatial resolution and the high noise of wireless signals.
For example, the range, azimuth, and elevation resolution of the device used in  \cite{zhao2019through} and  \cite{xue2021mmmesh} are 10 cm, 15 degrees, 15 degrees, and 4.3 cm, 15 degrees, 60 degrees, respectively.

Some works pioneer in the exploration of human body reconstruction from the wireless signals. Zhao et al.  \cite{zhao2018through} and Wang et al.  \cite{wang2019person} estimate 2D joint locations for humans. Zhao et al.  \cite{zhao2018rf} and Li~et al.  \cite{li2019making} push one step further, estimating the 3D joint locations. Full 3D human mesh reconstruction is first studied  \cite{zhao2019through}. As early works, these methods usually require customized equipments, which are cumbersome and hard to reproduce. Xue et al.  \cite{xue2021mmmesh} try to recover full 3D human body shapes and discriminate the gender with a commercial portable mmWave device. Despite the inspiring exploration, these works have not evaluated the accuracy quantitatively of reconstructing 3D human mesh from commercial mmWave radar devices in different scenarios and how they perform compared with RGB and depth cameras. Therefore, in this paper, we make efforts in answering the following questions. Can mmWave radars work robustly in different environments as claimed for 3D body reconstruction? Can they achieve comparable accuracy with RGB cameras or depth cameras? 

To answer these questions and push the 3D body reconstruction from wireless signals towards more diverse scenarios, we first design a data collection system with automatic 3D body mesh annotation, which is realized by fitting the SMPL-X body model \cite{SMPL-X:2019} to markers attached to subjects using MoSh++ \cite{mahmood2019amass}. Using this system, we collect a large-scale mmWave 3D human body dataset (denoted as mmBody) with 100 motions captured from 20 volunteers in 7 different scenes.
The statistics and visualization of the dataset in \cref{tab:Datasets compare} and \cref{fig:Compare_TSNE} reveal that our dataset makes a significant advancement in terms of completeness and diversity of scenarios, shapes, and poses. In addition to the mmWave signals, we also collect synchronized and calibrated RGB(D) images, which could open the possibility of research on combing mmWave radars with different sensors for the 3D body reconstruction.

With this dataset, we can then evaluate 3D body reconstruction performance in different scenarios using different sensor inputs. Specifically, models based on state-of-the-art learning-based methods using inputs from different sensors are trained on the dataset and tested in different scenarios. Metrics for different levels of reconstruction details are evaluated to show the capability of mmWave radar signals and compared to those from RGB(D) cameras.

In summary, the contributions of the paper are as follows:

\begin{itemize}
    \item An automatic annotation system for 3D human body reconstruction from mmWave radar signals is built, and a large-scale mmWave human body dataset with paired RGBD images is introduced.
    \item A benchmark deploying a state-of-the-art method for the reconstruction from mmWave signals is constructed and its performance across different scenes is evaluated.
    \item The reconstruction from mmWave signals vs. RGB(D) images in different scenarios is extensively compared to give quantitative evaluations about the capability of mmWave radars in 3D body reconstruction and guide the combination of signals from different sensors. 
\end{itemize}

\section{Related Work}
\label{sec:formatting}

\begin{table*}[ht]
\setlength{\belowcaptionskip}{0.2cm}
  \centering
  \resizebox{\textwidth}{!}{
  \begin{tabular}{ccccccccccc}
    \toprule
     &  &  & & &\multicolumn{5}{c}{Scenes}  \\ \cline{5-10}
    Dataset & Signals & Labels  &  No. Actions &Public & Occlusion & Poor Lighting & Furnished & Rain & Smoke \\
    \midrule
    RF-Pose  \cite{zhao2018through} & RF Signal&2D Skeletons  & / & \XSolid  &  \Checkmark & \Checkmark & \XSolid & \XSolid & \XSolid \\
    RF-Pose3D  \cite{zhao2018rf} & RF Signal& 3D Skeletons  &  /  & \XSolid  & \Checkmark & \XSolid & \XSolid & \XSolid & \XSolid \\
    RF-MMD  \cite{li2019making}& RF Signal& 3D Skeletons  &  35 & \XSolid  & \Checkmark & \Checkmark & \XSolid & \XSolid & \XSolid \\
    Person-in-WiFi  \cite{wang2019person}& Wi-Fi& 2D Skeletons  &  / & \XSolid & \XSolid & \XSolid & \XSolid & \XSolid  & \XSolid \\
    RF-Avatar  \cite{zhao2019through} & RF Signal& 3D Mesh & / & \XSolid & \Checkmark & \XSolid & \XSolid & \XSolid & \XSolid\\
    mmMesh  \cite{xue2021mmmesh} & mmWave & 3D Mesh  &  8 & \XSolid & \Checkmark & \Checkmark & \Checkmark & \XSolid & \XSolid \\
    Ours & mmWave, RGB(D) & 3D Skeletons/Mesh  &  100  &\Checkmark &\Checkmark & \Checkmark &\Checkmark & \Checkmark & \Checkmark \\
    \bottomrule
  \end{tabular}}
  \caption{Comparison of 3D body datasets with wireless signals}
  \label{tab:Datasets compare}
 \vspace{-0.5cm}
\end{table*}

\noindent\textbf{3D Human Body Reconstruction} Reconstruction from RGB(D) images has been researched for many years. Methods to solve the problem can be broadly divided into two categories: model-based methods and learning-based methods. In the former, a body model, e.g. SMPL  \cite{loper2015smpl}, is constructed to represent the human body, and an energy function measuring the difference between the body model and the input is optimized to find the best model parameters  \cite{ajanohoun2021multi,huang2017towards,bogo2016keep}. The methods require no labeling and can generalize to unseen data, while the optimization usually cannot achieve real-time performance due to the large search space. In the latter, a mapping function from the input to the output representation of the body, e.g. 2D/3D skeletons  \cite{sun2018integral, pavlakos2017coarse,martinez2017simple, wang2018drpose3d,chen20173d}, the parameters of SMPL  \cite{kanazawa2018end, kolotouros2019learning, kocabas2020vibe, pavlakos2018learning, lassner2017unite} is learned. Though efficient, the methods are data-hungry. In our work, we use both methods, model-based methods for getting annotations from MoCap inputs and learning-based methods for the real-time reconstruction of the 3D body.

\noindent\textbf{Wireless-based Human Sensing} Several wireless systems have been developed to reconstruct the human body  \cite{zhao2018through,zhao2018rf,li2019making,wang2019person,zhao2019through,xue2021mmmesh} and the mmWave-based system is one of them. The mmWave sensing has been widely adopted to enable various human sensing works, such as human monitoring and tracking \cite{zeng2016human,alizadeh2019remote,wang2020remote}, human detection and identification \cite{gu2019mmsense,yang2020mu}, and behavior recognition \cite{zhang2018real,smith2018gesture,kwon2019hands}. For human pose estimation, only a few works \cite{sengupta2020nlp,li2020capturing,sengupta2020mm} have been presented recently. In \cite{sengupta2020nlp}, the authors present a human pose estimation model using a natural language processing approach based on simulated mmWave radar point clouds 
and evaluate the model in normal scenes. 
Further, Li \etal \cite{li2020capturing} present an accurate human pose estimation system that relied on a 77GHz mmWave radar, and Sengupta \etal \cite{sengupta2020mm} propose a real-time human skeletons detection and tracking approach using a mmWave radar. 
Works on the full-body reconstruction from mmWave signals are limited, \ie  the seminal  work \cite{zhao2019through} and a very recent one  \cite{xue2021mmmesh}. Zhao \etal \cite{zhao2019through} reconstruct the 3D
human mesh by utilizing RF signals to build up a hardware system, which demonstrates that wireless signals contain sufficient information for the estimation of the pose and shape of the human body. To make the reconstruction more accessible, Xue \etal \cite{xue2021mmmesh} present a real-time human mesh estimation system using commercial portable mmWave devices. However, the datasets are not public in these works and the capability of the reconstruction from the mmWave signals in adverse environments is not explored. Its performance compared with the reconstruction from RGB(D) images is not studied either.

\noindent\textbf{mmWave-based Datasets} With the demand for autonomous driving, some mmWave-based datasets \cite{wang2021rethinking, bijelic2020seeing,caesar2020nuscenes} have been proposed recently for object detection and semantic understanding, while no public mmWave-based datasets for 3D body reconstruction are available, which limits the development of this field to some extent. 
Our annotation system and dataset aim to fill the gap. 
In addition, our captured data enable us to compare the performance of mmWave radars with RGB(D) cameras quantitatively, providing valuable guidance for the design of the mmWave radar sensing system and the combination of mmWave radars and cameras.

\begin{figure}[t]
  \centering
  \includegraphics[width=0.85\linewidth]{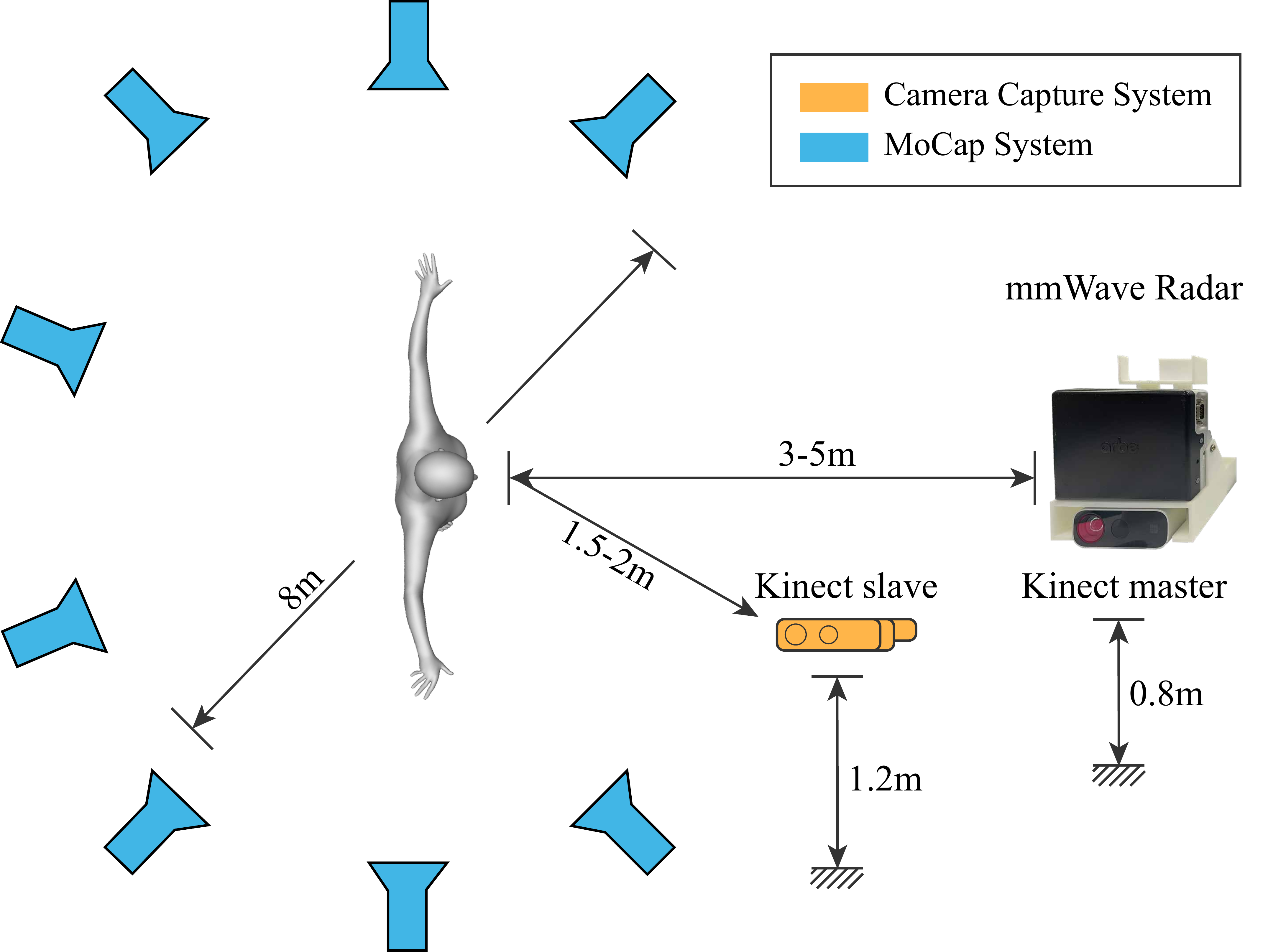}
  \caption{Hardware System}
  \label{fig:Hardware}
\end{figure}

\section{Full body Annotation }

\seclabel{annotation}
In this section, we present our method to build an automatic annotation system. The hardware system consists of three parts: a mmWave capture system to record the body dynamics, a MoCap system to label both human joint locations and full-body meshes, and a camera capture system to obtain RGB(D) images. The spatial arrangement of each component of the hardware system is roughly set as shown in \cref{fig:Hardware}. The hardware system is explained in \secref{hardware}. Then the synchronization and calibration among the systems follow in \secref{calibration}. In \secref{fitting}, the acquisition of the full-body mesh annotation is discussed. \secref{benchmark} shows statistics of our benchmark. \secref{comparison} gives a comparison of pose and shape space of mmBody with popular human body datasets captured using MoCap or RGB(D) images.

\subsection{Hardware System}
\seclabel{hardware}

\noindent \textbf{mmWave System} 
The spatial sensing via mmWave is achieved by transmitting wireless signals and receiving their reflections from environments via antenna arrays. The frequency shift between transmitting and receiving signals and the difference in arriving time between antennas determine the range measurement and the angle measurement, respectively. In addition to the spatial sensing, mmWave can measure the range velocity via the Doppler effect. For more details on the mmWave spatial sensing mechanism, we refer readers to the technical report \cite{1843-website}.

In our work, we choose the Phoenix type mmWave radar produced by Arbe Robotics \cite{Arbe-Robotics-website} for its high resolution, which works at 10 to 30 FPS. An antenna array of 48 transmitting channels by 48 receiving channels enables it to reach 0.4 meters for the range resolution and about 2.0 degrees for the angle resolution. It has an onboard processor to convert the original signals into point clouds which we use as input for 3D body reconstruction. 
More specifications of the radar are provided in the product overview \cite{radar-overview}.

The mmWave radar is placed on a 3D printing holder with a depth camera (Azure Kinect \cite{Kinect-website}) beneath it, shown in  \cref{fig:Hardware}. The holder is fixed on a tripod about 0.8 meters above the ground. The dynamics of subjects are captured at a distance of 3 to 5 meters away from the radar. The mmWave radar captures the scene at about 14 FPS. Each point of the generated point cloud for a frame contains its 3D location, range velocity, amplitude, and energy power of a reflected wave of the corresponding point in the scene.

\noindent \textbf{Camera Capture System} The camera capture system aims to get the RGB(D) images. The system consists of 2 Azure Kinects: a master Kinect is placed right under the mmWave radar, and the other slave one is located on one side of the radar-body line, 1.5-2 m from the body. As the mmWave radar works well at a distance of 3 meters away and the quality of the depth images degrades with distance, we place the slave Kinect closer to the subject to ensure good depth quality and thus a fair comparison with the radar. The Kinects are connected using synclines. Azure Kinects provide color images and depth images at a speed of 30 FPS. 

\noindent \textbf{MoCap System} The MoCap system aims to provide the 3D body skeletons and full-body meshes. It is the main annotation system for our dataset collection. Our OptiTrack\cite{OptiTrack-website} MoCap system consists of 8 cameras and markers attached to the human body. Cameras are evenly fixed on 8 tripods around a circular field with a radius of 8 meters at the height of 2.5 m, all looking at the center of the field. The system provides high-quality maker locations (accuracy of 0.8 mm) at a speed of 300 FPS at most. The number of markers is 37 and most markers are attached near human joints.

\subsection{Calibration and Synchronization}
\seclabel{calibration}

\noindent \textbf{Calibration} We set the mmWave radar coordinate frame as the target coordinate frame and transform the labels obtained from the MoCap system and the camera capture system to it.

The calibration between the mmWave radar and the camera capture system is achieved in two steps. The first step is the calibration of the Azure Kinect sensors. It is calibrated using a 1m $\times$ 1m Aruco tag, and the transformation matrix is obtained via the Colored ICP algorithm \cite{park2017colored}. The second step is the calibration between the mmWave radar and one of the Azure Kinect sensors. Following \cite{lu2020milliego} this, we place the mmWave radar and the sensor on a 3D printing holder. The transformation matrix between the two sensors is set beforehand. The calibration between the mmWave radar and the MoCap system is achieved by placing markers on the radar and using the position of markers located by the OptiTrack system to calculate the transformation matrix. 

\noindent \textbf{Synchronization} 
As the three systems work at different operation systems, synchronization is needed. The synchronization between the mmWave system and the camera capture system is done by running the mmWave system on the Ubuntu VM of the physical machine where the camera capture system running on. For the MoCap system (running on another PC), we synchronize it to the camera capture system via local network connection. For the mmWave radar, we can only get the timestamp of receiving the point clouds and therefore, are not able to get the exact capture time between the timestamps for two frames. The miss-alignment between the mmWave radar and other data is manually checked and adjusted slightly for each sequence. 

\subsection{Full Body Annotation}
\seclabel{fitting}

To obtain full-body mesh annotations, we use MoSh++ \cite{mahmood2019amass} to fit the parameterized body representation, i.e. SMPL-X \cite{SMPL-X:2019} to marker locations from the MoCap system. The SMPL-X model is defined as a function $\mesh(\shape, \pose, \trans)$, where $\shape$ represents shape parameters, $\pose$ body pose, hand pose and facial expression parameters,  and $\trans$ translation. For the body pose parameters, the first 3 dimensions represent the global rotation of joints, and the rest represent rotations of 21 body joints. 
For the hand pose and expression parameters, we use their template values and keep them fixed. We leave the recovery for these detailed 3D structures as future work. In the following paper, we use $\pose$ for body pose only. The output of the function $\mesh(\shape, \pose, \trans)$ is a triangulated mesh. The second row of \cref{fig:GoodCase} shows the meshes are aligned well with RGB images and radar point clouds.


\begin{figure*}
    \centering
    \includegraphics[width=0.85\linewidth]{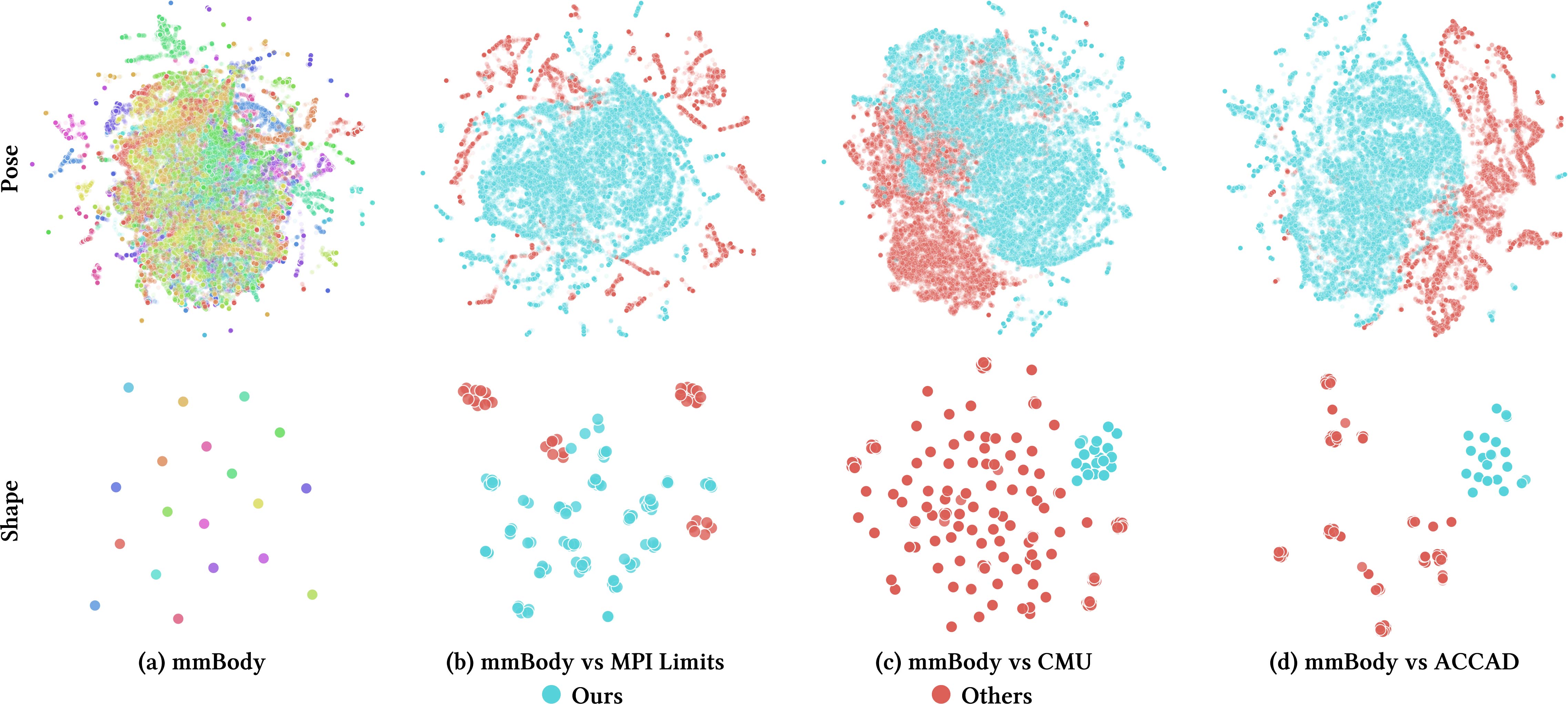}
    \captionof{figure}{2D-TSNE embedding of poses and shapes of mmBody and other datasets. The color of the dots in (a) represents different subjects.}
    \label{fig:Compare_TSNE}
\end{figure*}

\subsection{Build a Complete and Concise Benchmark}
\seclabel{benchmark}

We collected more than 200k frames covering 100 motions of 20 volunteers in 6 different environments. Among the 100 motions, there are 16 static poses, 9 torso motions, 20 leg motions, 25 arm motions, 3 neck motions, 14 sports motions, 7 daily indoor motions, and 6 kitchen motions. Among the 20 volunteers, there are 10 females and 10 males (physical gender), with weights ranging from 42kg to 75kg and heights ranging from 159cm to 183cm. 
\cref{fig:Compare_TSNE} (a) shows the TSNE visualization of our SMPL-X body shape space and pose space, demonstrating the completeness and evenness of the pose and shape coverage of our dataset.



The 6 different scenes include 2 different labs, a furnished lab, poor lighting, rain, smoke, and occlusion. We record occlusion scenes with mmWave radar covered with different materials (paper, plastic wrapping paper, foam board, cloth) to test its penetration ability. In the furnished scene, the furniture is randomly placed behind the human activity area in the lab. For the smoke and rain scene, we simulate smoke/fog weather using smoke cakes and rain weather using a showerhead. \cref{fig:GoodCase} shows different environments and ground truth meshes for the subjects in the scenes.


\subsection{Comparison with Other Datasets}
\seclabel{comparison}

To show the coverage of our dataset better, we compare the pose space and the shape space of mmBody with three popular datasets for human body reconstruction using MoCap or RGB(D) images, \ie the CMU dataset \cite{CMU}, the MPI Limits, \cite{akhterPoseConditionedJointAngle2015} and the ACCAD dataset \cite{ACCAD}. The comparison of the 2D TSNE of SMPL-X poses and shapes of these datasets is shown in \cref{fig:Compare_TSNE}. The SMPL-X parameters of the other three datasets are from the AMASS \cite{mahmoodAMASSArchiveMotion2019}, a large and varied database of human motion. The AMASS provides the SMPL+H \cite{romeroEmbodiedHandsModeling2017}, we only use the first 21 joints of SMPL+H, which are identical to SMPL-X.

\cref{fig:Compare_TSNE} (b) compares the pose and shape space with the MPI Limits dataset \cite{akhterPoseConditionedJointAngle2015} (referred to as PosePrior in AMASS). The AMASS provides 35 motions of the MPI Limits, at a total length of 20.82 minutes. This dataset aims to model the pose priors over 3D human pose and the subjects are instructed to perform extreme poses. The TSNE embedding reflects the extent of these limits that mmBody fails to reach but mmBody covers a very even space within these limits. The shape space of mmBody has a border coverage. \cref{fig:Compare_TSNE} (c) compares the pose and shape space with the CMU dataset \cite{CMU}. The CMU MoCap dataset contains 2605 trials in 6 categories and 23 subcategories. The AMASS provides SMPL-X parameters containing 2083 motions of 106 subjects, at a total length of 551.56 minutes. Though our dataset only consists of 100 motions, about 5\% of the CMU motions, the pose space covers a similar large space. \cref{fig:Compare_TSNE} (d) compares the pose and shape space with the ACCAD dataset \cite{ACCAD}. The AMASS provides 252 motions of 20 subjects of the ACCAD, at a total length of 26.74 minutes. The ACCAD contains daily motions which mmBody covers, and stage actions like dance and performance which mmBody does not cover. The shape space is larger than mmBody.

\section{3D Body Reconstruction}
\subsection{Dataset Preprocess} 
Many methods for body pose estimation or mesh recovery require to have as an input the region of interest containing only the body part. To crop bodies, we use the bounding boxes automatically annotated from the ground truth skeletons. As our current focus is reconstruction, we leave the detection of humans from mmWave signals as future work. The quality of human detection is likely to affect body reconstruction.

Given a dataset $\mathcal{D}=\{\pcl_t, \smpl_t\}, t=0,...,N$, where $\pcl_t$ is the cropped body region of the input, either the mmWave radar point cloud, the Azure Kinect point cloud with/without RGB information or the RGB image at time $t$, and $\smpl_t$ is the annotation acquired in \secref{fitting}, i.e. SMPL-X parameters for the body $(\shape, \pose, \trans)$. 3D joint locations and the vertices of the human body mesh can be obtained by $\cajjoint=\joint(\smpl_t)$ and $\cajvertices=\mesh(\smpl_t)$ with ${\cajjoint}\in \R^{ 22\times 3}$ for 22 joints and ${\cajvertices}\in \R ^{10475\times 3}$ for 10475 vertices.

\subsection{Reconstruction}

\noindent\textbf{Reconstruction with Point Clouds} We choose P4Transformer \cite{fan2021point} for the point-cloud-based reconstruction as the method combines spatial and temporal information from input frames and shows the state-of-the-art accuracy in the tasks of point cloud classification and segmentation. For the estimation for time $t$, previous frames are used, and we abuse $\pcl$ a bit to represent both a single frame and several frames. Our target is to learn a mapping $\smpl=f(\pcl)$, or $(\shape, \pose, \trans)=f(\pcl)$. 

For a point belonging to the Azure Kinect point cloud, its input features include the 3D coordinates of the point or its corresponding RGB values. For a point belonging to the radar point cloud, its input features include its 3D coordinates,  amplitude,  velocity, and energy.

The pose parameters of $\smpl$ are represented as 3D axis-angle vectors in SMPL-X which are not continuous in the real Euclidean space of four dimensions and thus hard for the neural network to learn according to  \cite{zhou2019continuity}. Therefore, we use the 6D representation proposed in the paper to represent the rotation. 

The loss function for the training is 
\begin{equation}
\label{eq: loss}
  \loss=\loss_{\trans}+\loss_{\pose}+\loss_{\shape} +\loss_{\jointloc} +\loss_{\vertloc},
\end{equation}
where $\loss_{\trans}, \loss_{\shape}, \loss_{\jointloc}, \loss_{\vertloc}$ is L1 loss between the prediction and the ground truth for translation, SMPL-X shape parameters, joint locations and vertex locations, and $\loss_{\pose}$ geodesic loss for SMPL-X pose parameters  \cite{mahendran20173d}. 
The geodesic loss is defined as the distance between rotation matrices, \ie 
$\cos ^{-1}[\frac{\operatorname{tr}\left(R_1R_2^T\right)-1}{2}]$.

\noindent\textbf{Reconstruction with RGB Images} We take the state-of-the-art method, VIBE~\cite{kocabas2020vibe}, to verify the accuracy of human pose and shape reconstruction with RGB images. VIBE estimates the SMPL-X body model for each frame using a temporal generation network, which is trained together with a motion discriminator. This discriminator has access to a large corpus of human motions in SMPL-X format.
The model firstly crops the human area from the images using the prior information of the 2D bounding box. Then it converts the size of the crop into 224$\times$224 as the input of the network. The output of the network is SMPL-X parameters for each frame. To train the network, it calculates 3D joints, pose, shape, and adversarial loss between prediction and label.




\begin{figure*}[h!]
    \centering
    \includegraphics[width=0.83\linewidth]{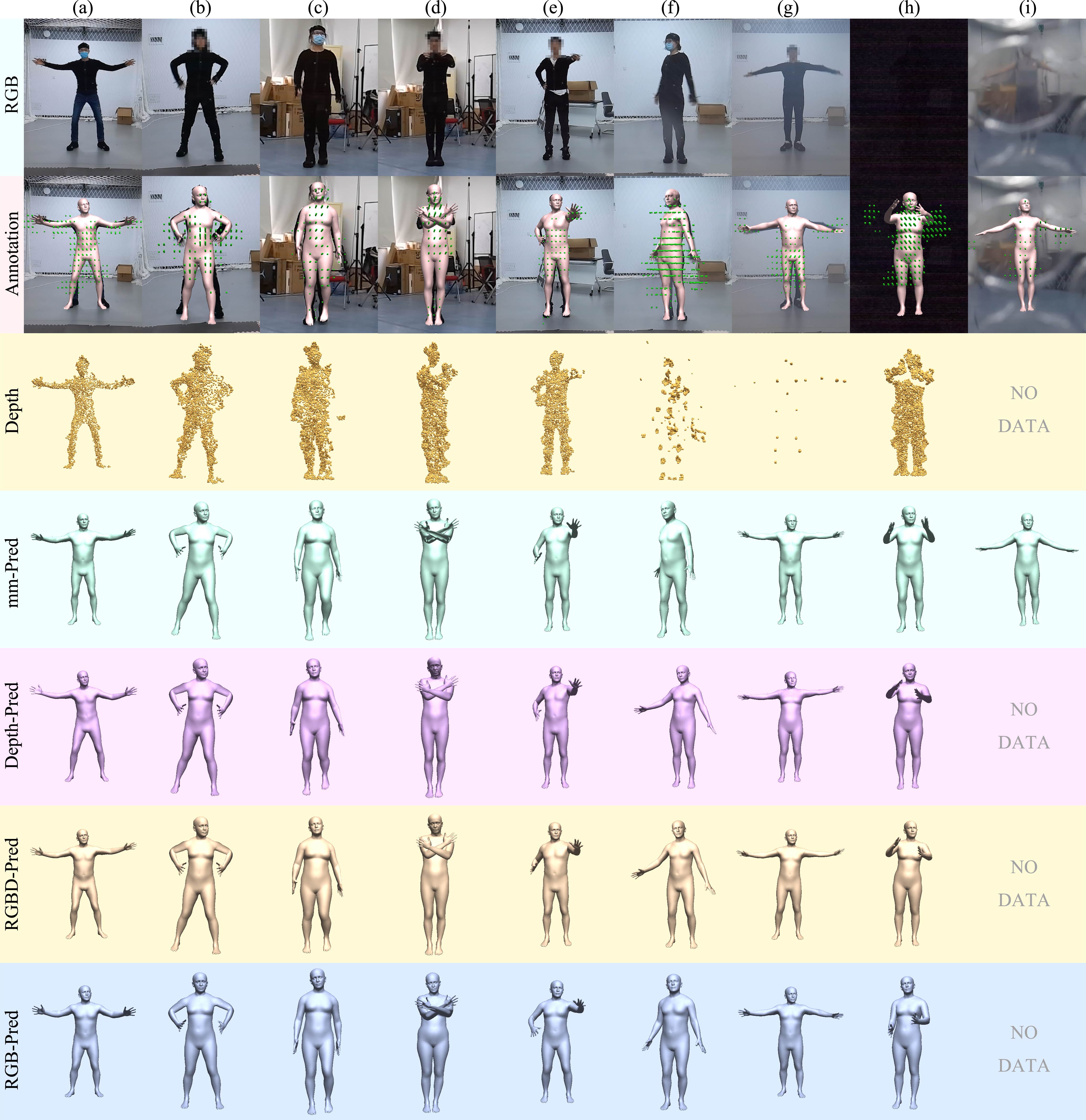}
    \captionof{figure}{Qualitative results. Column (a) (b), (c) (d), (e), (f), (g), (h), (i) show the different model inputs and reconstruction results in the lab1, lab2, furnished, rain, smoke, poor lighting, and occlusion scene, respectively. }
    \label{fig:GoodCase}
\end{figure*}

\section{Experiment and Analysis}

\noindent\textbf{Metrics and Dataset Split} To evaluate the performance of the reconstruction, we use  mean (max) joint error per frame and mean (max) vertex error per frame, which measures the average (maximum) Euclidean distance between the prediction and the ground truth for joints/vertices in each frame. The max joint/vertex error is more strict than the mean joint/vertex error. For the max vertex error, we also draw the curves for the proportion of frames with the max vertex error for each frame under different thresholds (use the max vertex error curves for short). As the metrics on the vertex error entangle the errors resulting from the pose and the shape, the mean squared error between the ground truth shape and the estimated one is adopted to evaluate the shape estimation using different inputs.

\begin{table}[H]
\setlength{\belowcaptionskip}{0.2cm}
\resizebox{\linewidth}{!}{
  \centering
  \setlength{\belowcaptionskip}{-3mm}
    \begin{tabular}{cccccccc}
    \toprule
    Scenes & Lab1 & Lab2 & Furnished & Poor Lighting & Rain & Smoke & Occlusion \\
    \midrule
    train & 10/4 & 10/6 & / & / & / & / & / \\
    test  & 2/2 & 2/2 & 2/2 & 2/2 & 2/2 & 2/2 & 2/2 \\
    \bottomrule
    \end{tabular}}
\caption{Training set and testing set. */* denotes the number of sequences/number of subjects.}
 \label{tab:datasetsplit}%
 \vspace{-0.8cm}
\end{table}%

The dataset is split into training and testing sets as Tab.\ref{tab:datasetsplit} shows. 
We choose 20 sequences from 10 subjects recorded in the lab scenes as the training set. For the testing set, except Lab2 containing seen subjects in the training, 2 sequences from different subjects for each scene are held out. Each sequence contains about 2000 frames of data.

\noindent\textbf{Experiment Setting}
To compare the reconstruction from the mmWave radar to that from RGBD images, we train three networks for point clouds based on the P4Transformer, the inputs being the radar point cloud, the point cloud from the Kinect, and the same point cloud with calibrated RGB features. The number of the radar points sampled from each frame for the network is 1024, and the number of the Kinect points is 4096.  The P4Transformer is implemented using Pytorch and is trained on an Nvidia GeForce RTX 3090. We train the entire network for 50 epochs from scratch with an Adam optimizer and an initial learning rate of 0.001. Other parameters are set to default. In terms of the RGB-based model, we train the VIBE \cite{kocabas2020vibe} with the input of 224$\times$224 images cropped from the RGB images. The epochs, optimizer, and initial learning rate are the same with P4Transformer and other parameters are set according to the VIBE. Training the VIBE to converge on one Nvidia GeForce RTX 3090 takes around 3 hours.

\subsection{3D body reconstruction from mmWave} 
The point clouds generated by the mmWave radar are usually very sparse, and contain many missing parts and noise resulting from the multi-path effect. Particularly, with such low-resolution point clouds, its ability to reconstruct the full 3D body is questioned.

\noindent \textbf{Joint and Vertex Error} Our experiment results show that the 3D body can be reconstructed well from the mmWave radar signals in spite of the sparsity. The mean joint error and the mean vertex error can reach as low as 8cm and 10cm. The reconstructed meshes from mmWave radar point clouds for different poses and subjects in the different scenarios are shown in \cref{fig:GoodCase}. Overall, the reconstructed meshes for most samples are close to the ground truth.

\begin{figure*}[h]
\centering
\setlength{\abovecaptionskip}{-0.2mm}
\setlength{\belowcaptionskip}{-4mm}
    \includegraphics[trim=200mm 0mm 200mm 0mm,clip,width=0.9\linewidth]{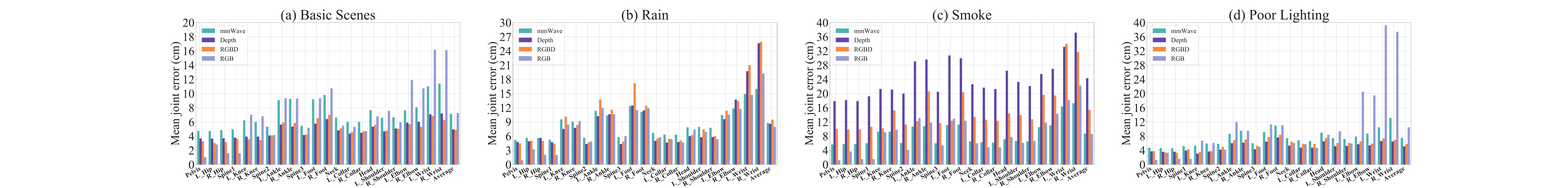}
   \caption{Mean joint errors of each body joint for the mmWave radar and the RGB(D) camera in different scenes.}
    \label{fig:PJPE}
\end{figure*}

\begin{figure*}[h]
\centering
\setlength{\abovecaptionskip}{-0.01mm}
\setlength{\belowcaptionskip}{-2mm}
    \includegraphics[trim=200mm 0mm 200mm 0mm,clip,width=0.9\linewidth]{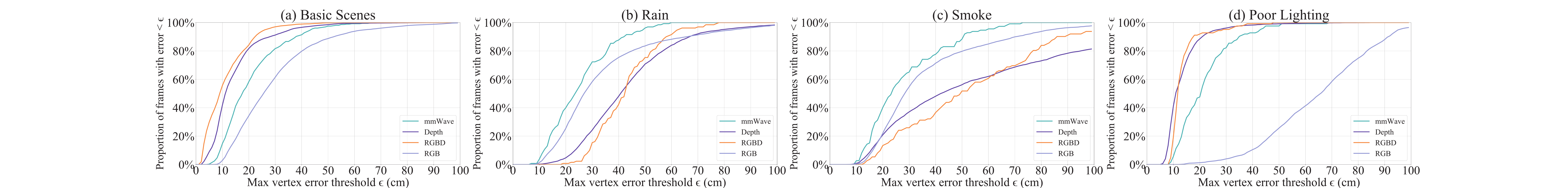}
    \caption{The proportion of frames with the max vertex errors within thresholds for different inputs in different scenes.}
    \label{fig:max_error_cdf}
\end{figure*}

The reconstruction from mmWave radar point clouds can achieve better accuracy than that from RGB images in the normal scenes. Though in \cref{fig:PJPE}, the average joint errors for the mmWave radar signal and RGB images are both 7cm, the max vertex error curve in \cref{fig:max_error_cdf} (a) shows the mmWave radar performs much better\footnote{Though we use different networks for the mmWave radar point clouds and the RGB images and the comparison is not strictly fair, we think the results of the state-of-the-art methods can serve as a reference.}. However, there exists a gap in the reconstruction accuracy between the mmWave radar and the Kinect depth sensor: 2cm more in the mean joint error and 20\% fewer frames whose max vertex errors are under 20cm for the reconstruction using the mmWave radar compared to the depth sensor. Adding RGB features into the Kinect point clouds improves the reconstruction from depth slightly.

\begin{figure*}[h]
\centering
\setlength{\abovecaptionskip}{-0.2mm}
\setlength{\belowcaptionskip}{-4mm}
    \includegraphics[trim=200mm 0mm 200mm 0mm,clip,width=0.9\linewidth]{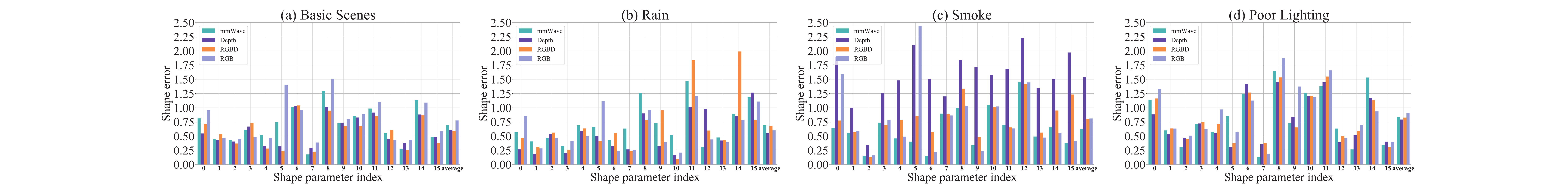}
   \caption{Shape errors for the mmWave radar and the RGB(D) camera in different scenes. }
    \label{fig:shape}
\end{figure*}

\begin{table*}[h]\footnotesize
\setlength{\belowcaptionskip}{0.2cm}
\resizebox{\textwidth}{!}{
  \centering
    \begin{tabular}{c|c|cc|cc|cc|cc|cc|cc|cc|cc}
    \toprule
    \multicolumn{2}{c|}{\multirow{2}[2]{*}{Scenes}} & \multicolumn{6}{c|}{Basic Scenes}             & \multicolumn{8}{c|}{Adverse Environments}                     & \multicolumn{2}{c}{\multirow{2}[2]{*}{Average}} \\
    \multicolumn{2}{c|}{} & \multicolumn{2}{c}{\textbf{Lab1}} & \multicolumn{2}{c}{\textbf{Lab2}} & \multicolumn{2}{c|}{\textbf{Furnished}} & \multicolumn{2}{c}{\textbf{Rain}} & \multicolumn{2}{c}{\textbf{Smoke}} & \multicolumn{2}{c}{\textbf{Poor Lighting}} & \multicolumn{2}{c|}{\textbf{Occlusion}} & \multicolumn{2}{c}{} \\
    \midrule
    \multirow{4}[1]{*}{Mean Error} & mmWave & 7.8   & 9.5   & 5.8   & 6.6   & 8.2   & 10.4  & 8.8   & 10.2  & 8.7   & 10.0  & 7.5   & 9.5   & 10.7  & 14.1  & 8.2   & 10.0 \\
          & Depth & 5.5   & 6.5   & 3.9   & 4.3   & 5.5   & 6.9   & 8.6   & 10.9  & 24.3  & 28.0  & 5.1   & 6.5   & /     & /     & 8.8   & 10.5 \\
          & RGBD  & 5.8   & 7.0   & 3.4   & 3.9   & 5.4   & 6.8   & 9.5   & 11.6  & 15.4  & 18.3  & 5.8   & 7.2   & /     & /     & 7.5   & 9.1 \\
          & RGB   & 7.4   & 8.9   & 7.3   & 10.0  & 7.1   & 9.1   & 8.0   & 10.1  & 8.6   & 10.8  & 10.5  & 15.6  & /     & /     & 8.1   & 10.8 \\
    \midrule
    \multirow{4}[1]{*}{Max Error} & mmWave & 16.9  & 22.5  & 13.3  & 18.8  & 17.5  & 25.5  & 20.0  & 26.3  & 20.5  & 29.0  & 16.2  & 22.6  & 25.3  & 35.3  & 18.5  & 25.7 \\
          & Depth & 12.6  & 17.2  & 8.8   & 12.7  & 11.3  & 16.4  & 29.8  & 44.6  & 49.4  & 61.7  & 10.3  & 14.4  & /     & /     & 20.3  & 27.8 \\
          & RGBD  & 12.2  & 16.5  & 7.5   & 10.9  & 10.1  & 14.1  & 29.0  & 43.7  & 38.8  & 53.4  & 11.2  & 14.5  & /     & /     & 18.1  & 25.5 \\
          & RGB   & 22.0  & 28.8  & 24.8  & 35.3  & 20.0  & 27.9  & 26.3  & 34.8  & 28.1  & 37.1  & 46.2  & 66.0  & /     & /     & 27.9  & 38.3 \\
    \bottomrule
    \end{tabular}%
    }
    \caption{Errors (cm) of 3D body reconstruction from the mmWave radar and the RGB(D) camera in different scenes. For the two columns of each scene, the first column is for joint error and the second vertex error.}
    \label{tab:sceneerror}
    \vspace{-1em}
\end{table*}%

\noindent \textbf{Shape Error} In the basic scenes, similar to joint and vertex errors, the reconstruction from the mmWave radar gives better shapes than that from the RGB images and worse than that from the depth images, shown in \cref{fig:shape} (a). In the adverse environments, the average shape errors for different inputs are close, except for the error for depth in the smoke scene, shown in \cref{fig:shape} (b-d).

\begin{figure}[h]
\centering
    \includegraphics[trim=0mm 0mm 0mm 0mm,clip,width=0.9\linewidth]{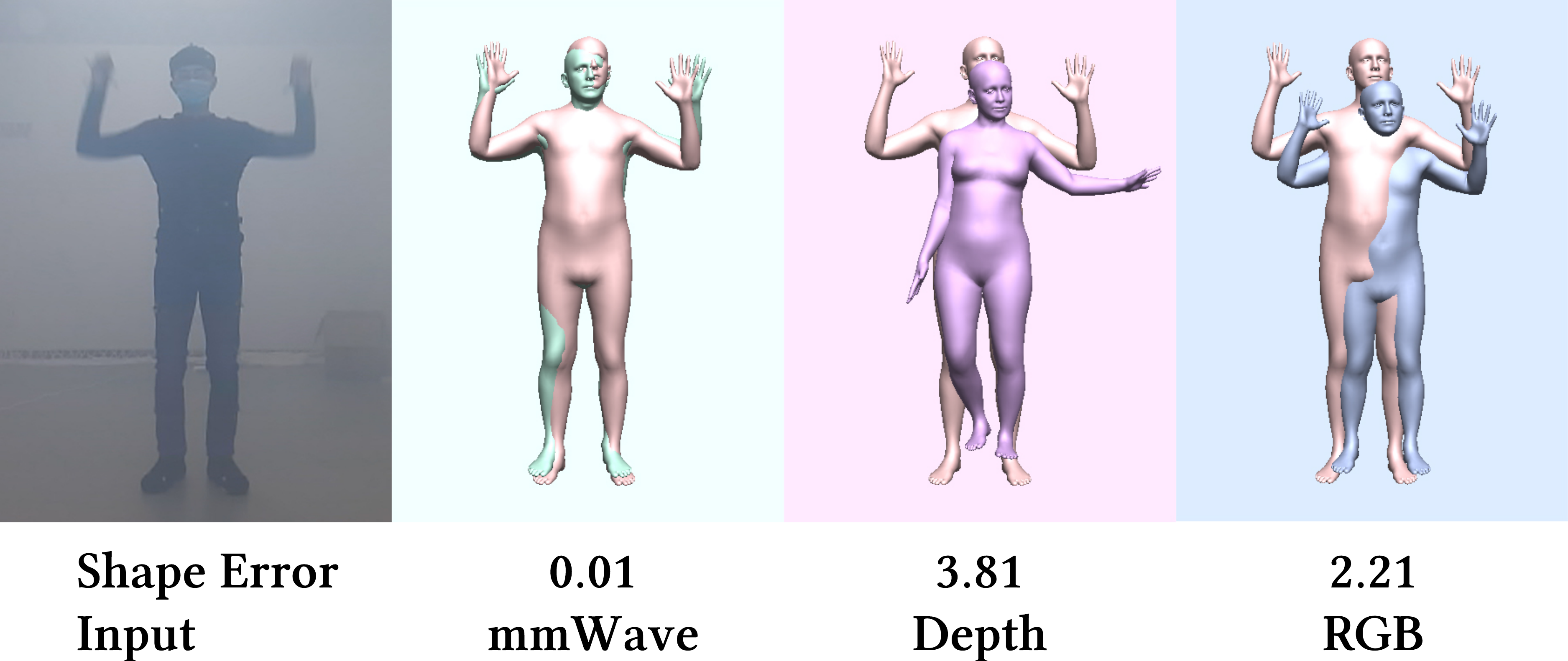}
    \caption{Comparison of shape errors for different inputs.}
    \label{fig:shape_comp}
    \vspace{-0.1cm}
\end{figure}

Though the average shape errors are close, the errors for different items of the shape vector can vary dramatically. In the smoke scene, for example, the average shape error for the mmWave radar and the RGB images are close (0.62 vs 0.75) while the error for the first shape item for RGB images triples that for the mmWave radar. Items in the shape vector control different aspects of the final shape. 
Deformation in only one item can cause large visual differences. \cref{fig:shape_comp} visualizes the estimation from different inputs. The text lists the errors for the first item of the shape vector.

\subsection{Robustness in different environments}
The main factors that prevent the generalization of the mmWave radar reconstruction to different scenes can be attributed to the multi-path effect noise and the absorption of energy in different spaces, like rain, and fog. 


The errors for the lab scenes without and with furniture (the columns for \textbf{Lab 1} and \textbf{Furnished} in \cref{tab:sceneerror}) show that the multi-path effect affects the performance of the reconstruction slightly. For example, the max joint error raises from 16.9cm to 17.5cm, and the max vertex error from 22.5cm to 25.5cm. 

Also, adverse environments like rain and smoke interfere the reconstruction moderately, the max joint errors in these environments increasing by 28\% (16.9cm to 20.0cm and 20.5cm) from that for the same scene without rain and smoke. For reference, the max joint error for the reconstruction from depth in the rain scene increases by 130\% (12.6cm to 29.8cm) and the reconstruction in the smoke scene almost fails, its max joint error reaching about 50cm. For the reconstruction in these environments, as all subjects dress in black and in the scenes the subjects can be clearly seen in the images, the errors only increase by 20\% and 27\%. Both the mmWave radar and the depth sensor are robust to poor lighting as they use active lighting while the RGB camera fails. However, occlusions cause the reconstruction performance to deteriorate by a large margin, the max vertex error increasing by 56\%.

\subsection{Challenges of the mmBody dataset}
\begin{figure}
\centering
    \includegraphics[width=0.8\linewidth]{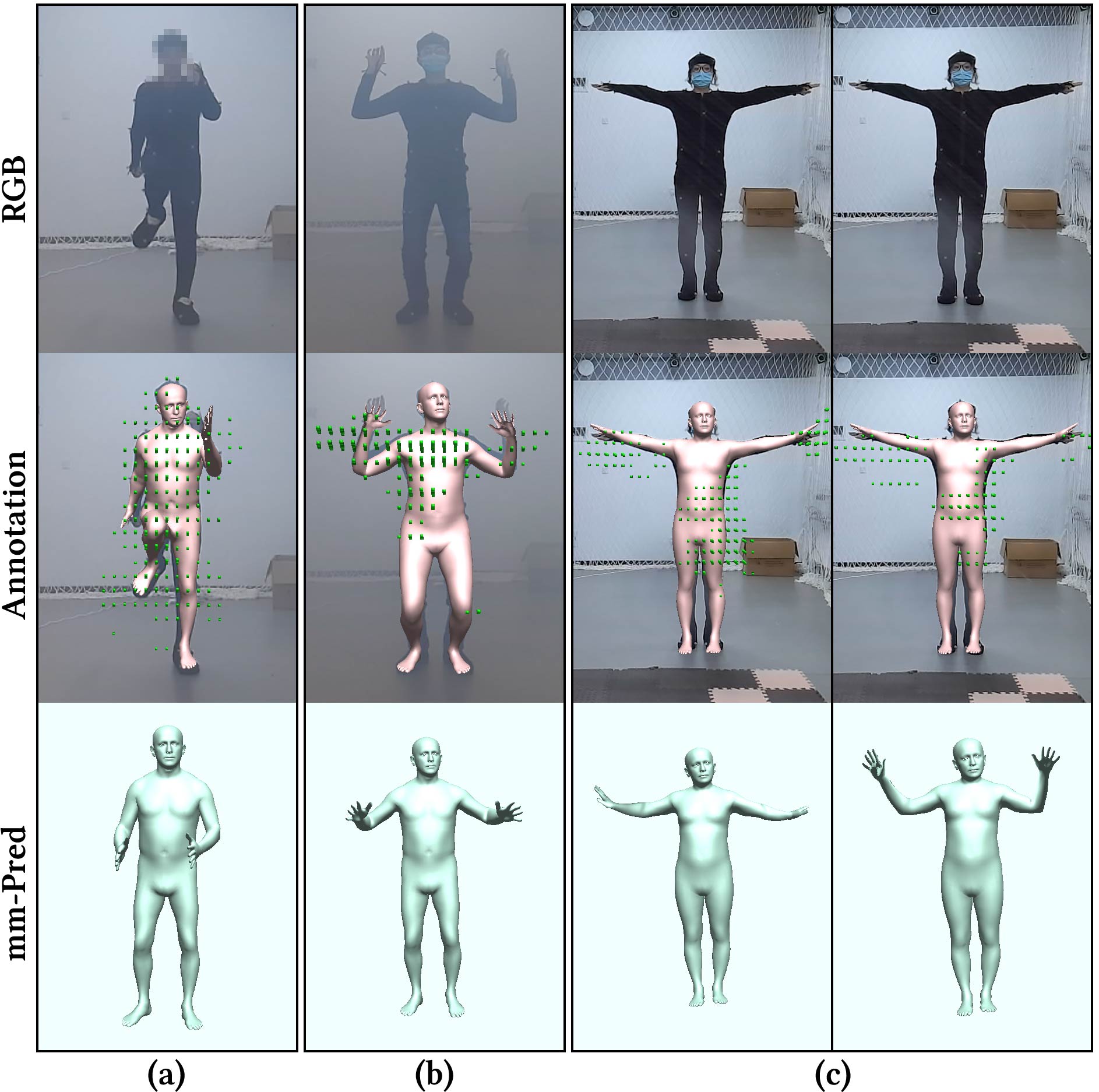}
   \caption{Challenges of our dataset.}
    \label{fig:challenges}
\end{figure}

There are situations when the reconstruction from the mmWave radar fails, some of which are exampled in \cref{fig:challenges}. The reasons for these failures can be attributed to sparse point clouds, large missing parts, and inconsistent radar point clouds. (1) Sparsity of radar points clouds (see the input radar point clouds in \cref{fig:challenges}): each frame of the mmWave radar only contains about 1k human points (at a distance of 3-5 m) due to the bandwidth and antennas of the Arbe Phoenix radar while a depth image contains up to 200k. 
(2) Large missing parts: some parts of the human body, such as the head and limbs, may not have radar points due to the specularity of mmWave signals, as shown in \cref{fig:challenges} (a) (b). (3) Inconsistent radar point clouds: due to the specularity and the multi-path effect of mmWave signals, almost identical actions may produce very different point clouds (\cref{fig:challenges} (c)). These failure cases pose particular challenges different from point clouds from the Kinect, and more sophisticated algorithms are required to deal with these challenges.

\subsection{Analysis for further improvements}

The comparison between reconstructions using different inputs provides cues for the design choice of input modalities and networks to improve the results. In \cref{fig:PJPE} (a), though the average of all joint errors for different inputs is of similar value, the variance for different joint errors is different. For the reconstruction using RGB images, the errors for the joints close to the center are very low (about 2cm for joints near the hip and 4cm for joints in the spine) while the errors for joints located at the ends of the limbs are significantly high (about 12cm for elbows and 16cm from wrists). For the reconstruction using the mmWave radar and the depth image, the variance is much smaller. This suggests different inputs and networks may bias toward different aspects of the reconstruction. Therefore, in addition to using the point clouds to represent the mmWave radar signals, 2D images and using 2D CNN for the feature extraction from the images can be considered, \eg projecting the point clouds into 2D depth images or representing the mmWave signals by 2D range heatmaps and azimuth heatmaps. The features processed by 2D CNN and networks like PointNet \cite{qi2017pointnet} can be further combined and help to improve the accuracy of joints near the human center.

Many applications require algorithms to work robustly all day in all weather conditions. Using only one sensor can hardly achieve the goal as we can see from \cref{tab:sceneerror}. Therefore, merging signals from different sensors and exploiting their complementary strengths is a promising solution. With the calibrated and synchronized mmWave signals and RGBD images, our dataset opens the potential for the study.


The comparison of the mmWave radar vs the RGBD camera reveals opportunities and challenges in combining different sensor signals. In the basic scenes, using both RGB and depth information achieves better reconstruction results than using only one of them (columns of Basic Scenes in \cref{tab:sceneerror}). As the mmWave radar signals are also represented as point clouds, we can expect adding the RGB information to the signals can help to gain better accuracy. However, due to the sparsity and large missing part of the point clouds, the combination may require a careful design instead of projecting a point to its calibrated 2D image and querying the corresponding pixel directly.  

Notice that in the adverse environments, the combination is not able to achieve lower error, only improving the worse case (columns of Adverse Environments in \cref{tab:sceneerror}). This may result from no available combination of signals for these environments in the training. Collecting data for different environments and learning a network for a simple concatenation of different signals in a black-box manner can improve. However, collecting data covering all the corner cases is non-trivial. In addition to those efforts, exploiting the confidence of the estimation of each signal in different scenarios and designing a dynamic weighting strategy for different signals may reduce the need for data and work more robustly.

\section{Conclusion}
In this paper, we have presented a 3D body reconstruction dataset for mmWave radars to close the gap of no available public datasets to study the problem of reconstructing the 3D human body from the noisy and sparse mmWave signals in different scenes. An automatic capture and annotation system is built up with multiple sensors. Extensive experiments and analyses are conducted to evaluate the performance of 3D body reconstruction from the mmWave radar in different scenes. The performance is also compared against RGB(D) images. The results show that 1) despite the noisy and sparse signals, the 3D body can be well recovered from the mmWave radar signal, even emulating the reconstruction from RGB images in our testing scenes; 2) the reconstruction from the radar is affected slightly by adverse environments like rain and smoke; 3) the reconstruction from the radar performs worse than depth camera in normal scenes, however, its robustness in extreme scenes is much better. Further, based on the analyses of the results, challenges particular to the radar point clouds are pointed out and insights for further improvements and the combination with RGBD cameras are shared.


\begin{acks}

This work was supported in part by NSFC under Grants  62088101, 61790571, 62103372, and the Fundamental Research Funds for the Central Universities.

\end{acks}

\bibliographystyle{ACM-Reference-Format}
\bibliography{sample-base}

\end{document}


\title{Supplementary Materials for mmBody Benchmark: 3D Body Reconstruction Dataset and Analysis for Millimeter Wave Radar}

\maketitle
\appendix
\section{Specifications of the Arbe Radar}

\bibliographystyle{ACM-Reference-Format}
\bibliography{sample-base}